# Implementation of a self-developed model predictive control scheme for vehicle parking maneuvers

Gergő Ignéczi
*Vehicle Industry Research Center*
*Széchenyi István University*
Győr, Hungary
gfigneczi1@gmail.com

Ernő Horváth
*Vehicle Industry Research Center*
*Széchenyi István University*
Győr, Hungary
herno@ga.sze.hu

Dániel Pup
*Vehicle Industry Research Center*
*Széchenyi István University*
Győr, Hungary
pupd@ga.sze.hu

*Abstract*—In this paper a self-developed controller algorithm is presented with the goal of handling a basic parking maneuver. One of the biggest challenges of autonomous vehicle control is the right calibration and finding the right vehicle models for the given conditions. As a result of many other research, model predictive control (MPC) structures have started to become the most promising control technique. During our work we implemented an MPC function from white paper. Considering the low-speed conditions of a parking maneuver we use a kinematic bicycle model as the basis of the controller. The algorithm has two main inputs: a planned trajectory and the vehicle state feedback signals. The controller is realized as a Simulink model, and it is integrated into a complete autonomous control system using ROS framework. The results are validated through multiple steps: using Simulink only with a pure kinematic bicycle plant model; using LGSVL simulation framework containing a real vehicle model and the entire software chain; the controller is prepared for real vehicle tests.

## I. INTRODUCTION

### A. Problem motivation

The role of motion control functions for assisted and automated vehicles have grown significantly in the last years. With the increasing number of available driver assistance functions in commercial usage, engineers are seeking control solutions for vehicle motion which are accurate and robust at the same time. Path tracking problem is usually divided into two main problems: longitudinal and lateral control. Most well spread control solutions, such as pure-pursuit techniques or PID control schemes are used to provide steering angle control to follow the pre-planned path laterally, while longitudinal control is a separated function aiming to follow a planned velocity trajectory or distance. The biggest disadvantages of these control structures is the difficult-to-handle constraints of the vehicle kinematics and the actuator operating range. The calibration of PID gains are often time consuming. The lateral behavior of the vehicle is speed-variant, therefore speed dependent control gains must be used. Thus, the longitudinal and lateral motion cannot be handled within a single controller.

### B. Literature overview

A compact solution is provided by model-based controls, especially by model predictive control (MPC) schemes. A possible solution for low-speed overtaking scenarios have been presented in [1]. The linear bicycle model for lateral and longitudinal control has been used to design an MPC controller. The study showed that MPC was able to meet the kinematic requirements even when simulating the controller with a non-linear model of the vehicle. A multiple-input multiple-output scheme has been presented in [1]. In this case the dynamic model of the vehicle was used to design an MPC which is able to control vehicle movement at the limits of handling for obstacle evasion maneuvers. This controller utilizes one of the biggest advantages of model predictive controls: the combination of lateral and longitudinal controls. In paper [2] an MPC was used for specifically low speed parking maneuvers. In this case the focus was on handling system saturation and position constraints, while the vehicle dynamic effects are neglectable and a pure kinematic model is sufficient to be used due to the lower speed range. The article proved that MPC algorithm can perform better in all scenarios than PID controllers.

The algorithm has two main inputs: a planned trajectory and the vehicle state feedback signals. The trajectory planner is not part of the paper. This trajectory contains equidistant waypoints as X and Y coordinates in a global reference frame, an orientation and a longitudinal velocity reference for each waypoint. The model predictive control operates on the global frame. As a first stage, a preliminary function calculates the actual waypoint from the input set – the one that is closest to the vehicle. Then the relatively large resolution of the waypoint array on the input (approximately 0.5 – 1.0 m distance between neighboring points) is resampled with a third-order interpolation on four surrounding waypoints – one behind the ego vehicle and three ahead of the ego vehicle. The outputs of the controller are the road wheel angle and the longitudinal acceleration. The optimization problem of the MPC is defined as a cost function of the plant input amplitude and the output deviation from the reference. Constraints are applied on the plant input gradient and on their absolute values. The optimization problem is solved with Hildreth's iterative approach.

In this paper we would like to introduce a self-developed model predictive controller based on [4], specifically designed for low-speed vehicle maneuvers. The aim of the study is to present how a model predictive controller is built up and how accurate it is able to control the path of the vehicle. The controller performance is validated through simulation using a self-developed model of the vehicle within Simulink and also using LGSVL simulation framework. The activity is related to a complete autonomous system development at the Vehicle Research Center of Széchenyi Istvan University.

## II. PROBLEM FORMULATION

### A. Research context

The design of the controller presented in this paper is related to a complete autonomous system development. The developed system is realized in a commercial electric vehicle which is a Nissan Leaf model year 2014. The autonomous system is built using Robot Operating System (ROS) running on a NVIDIA Jetson Xavier computer. The complete model of the vehicle and the detection functions (such as lidars, cameras) have been simulated within the LGSVL simulation framework. LGSVL is compatible with the ROS system and hence the functionalities are easy to be deployed and tested during the development phase. In chapter V. the controller performance based on pre-defined paths is presented.

### B. Control problem formulation

The activity presented in this paper mainly concentrates on providing controller solutions for low-speed vehicle automated maneuvers, which are usually used for parking missions and parking place searching missions. These maneuvers may contain drive-off and stopping tasks, velocity control tasks and path following tasks by actuating the steering system. As the tire slip effects are fairly small at speeds under 10 m/s, only the kinematic model of the vehicle is used for the controller design. The input of the controller is a pre-planned trajectory in the form of position-based waypoints which serve as the target lateral displacement over distance, the target orientation of the vehicle and the target speed. The trajectory is provided in the global world coordinate frame, and as such can be considered constant partially or fully during the maneuver. The vehicle position feedback is provided by a superordinate localization function which mainly relies on DGPS coordinates. The output of the controller can be the target road wheel angle of the front wheels and the longitudinal acceleration or the target longitudinal velocity of the vehicle. Both solutions are presented. In the first case the velocity control is also done by the model predictive control itself, which results a compact multiple input – multiple output (MIMO) controller structure for longitudinal and lateral control at the same time. This is used when presenting simulation results from the self-developed Simulink framework. When the target velocity of the vehicle is the output of the controller the velocity control task is executed by a subordinate function at the low-level control layer. This is used in the case of our test vehicle, and as such also in the LGSVL simulation framework.

## III. VEHICLE MODEL

For the controller design the kinematic bicycle model of the vehicle is used. In our self-developed Simulink simulation environment, the same model is used for validation. The LGSVL simulation framework uses the full dynamic model of the vehicle. The controller performance for each model is presented in chapter V. The kinematic bicycle model relies on the basic mechanical consideration of the vehicle body described in [5]. The relations can be seen in Fig. 1.

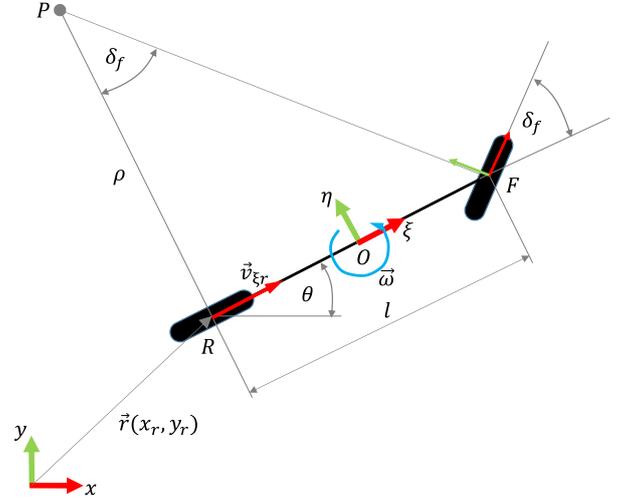

Fig. 1. Mechanical relations of the kinematic bicycle model

The kinematic equations forming the bicycle model are described by (1) – (6), while the discrete time state space model is defined in (7) – (12).

$$x_r = \int_0^t v_{\xi r} * \cos(\theta) d\tau + x_{r0} \quad (1)$$

$$y_r = \int_0^t v_{\xi r} * \sin(\theta) d\tau + y_{r0} \quad (2)$$

$$v_{\xi r} = \int_0^t a_{\xi r} d\tau + v_{\xi r0} \quad (3)$$

$$\theta = \int_0^t \omega \, d\tau + \theta_0 \quad (4)$$

$$\omega = \frac{v_{\xi r}}{\rho} = v_{\xi r} \kappa = v_{\xi r} \frac{1}{L} tg(\delta_f) \quad (5)$$

$$\frac{L}{\rho} = tg(\delta_f) = L\kappa \quad (6)$$

Where:

- $[x_r \; y_r]$ are the global position of the rear axle center point,
- $[x_{r0} \; y_{r0}]$ are the initial position of the rear axle center point in the global world frame,
- $v_{\xi r}$ is the longitudinal velocity of the vehicle, $v_{\xi r0}$ is the initial longitudinal velocity of the vehicle,
- $a_{\xi r}$ is the longitudinal acceleration of the vehicle,
- $\theta$ is heading angle of the vehicle in the global world frame, $\theta_0$ is the initial heading (orientation) of the vehicle,
- $\omega$ is the yawrate of the vehicle,
- $\rho$ is the radius, while $\kappa$ is the curvature of the path run by the rear axle center point,
- $L$ is the distance between the front and the rear axle,

- $\delta_f$ is the road-wheel angle of the front wheel.

From (1) – (6) the discrete state space model of the kinematic bicycle model can be derived as follows:

$$x_{k+1} = A(t)x_k + B(t)u_k \qquad (7)$$

$$y_{k+1} = Cx_{k+1} + Du_{k+1} \qquad (8)$$

Where:

$$x_k = \begin{bmatrix} x_r \\ y_r \\ v_{\xi r} \\ \theta \end{bmatrix} \qquad (9)$$

$$u_k = \begin{bmatrix} a_{\xi r} \\ tg(\delta_{f0}) \end{bmatrix} \qquad (10)$$

$$A(t) = \begin{bmatrix} 1 & 0 & T_s\cos(\theta_k) & 0 \\ 0 & 1 & T_s\sin(\theta_k) & 0 \\ 0 & 0 & 1 & 0 \\ 0 & 0 & 0 & 1 \end{bmatrix} \qquad (11)$$

$$B(t) = \begin{bmatrix} 0 & 0 \\ 0 & 0 \\ T_s & 0 \\ 0 & \frac{T_s}{L}v_{\xi r_k} \end{bmatrix} \qquad (12)$$

$$C = \begin{bmatrix} 0 & 1 & 0 & 0 \\ 0 & 0 & 1 & 0 \\ 0 & 0 & 0 & 1 \end{bmatrix} \qquad (13)$$

$$D = 0 \qquad (14)$$

### IV. Controller design

Based on the kinematic model presented in chapter III. the MPC can be designed. The design will be presented in two steps: first the unconstrained solution is introduced, then the optimization along input constraints is presented.

#### A. Unconstrained model predictive control

For the MPC design the so-called augmented model is introduced. This is differential state space model that can be directly derived from the plant model. The new state vector, input and output vectors are calculated as the delta vectors between the $k^{th}$ and the $(k+1)^{th}$ state of the system as written in (15) and (16). The vectors with upper waves are the vectors from the unchanged state space model.

$$\Delta\tilde{x}_{k+1} = \tilde{x}_{k+1} - \tilde{x}_k = A_d\Delta\tilde{x}_k + B_d\Delta\tilde{u}_k \qquad (15)$$

$$\Delta\tilde{y}_{k+1} = C_d(\tilde{x}_{k+1} - \tilde{x}_k) = C_d A_d \Delta\tilde{x}_k + C_d B_d \Delta\tilde{u}_k \qquad (16)$$

From equations (15) and (16) a new system can be defined, whose new state vector is defined in (17) and the state equations are defined in (18) – (19). This system is called the augmented model.

$$x_k = [\Delta\tilde{x}_k^T \ \tilde{y}_k]^T \qquad (17)$$

$$\begin{bmatrix} \Delta\tilde{x}_{k+1} \\ \tilde{y}_{k+1} \end{bmatrix} = \begin{bmatrix} A_d & o_m^T \\ C_d A_d & 1 \end{bmatrix}\begin{bmatrix} \Delta\tilde{x}_k \\ \tilde{y}_k \end{bmatrix} + \begin{bmatrix} B_d \\ C_d B_d \end{bmatrix}\Delta\tilde{u}_k \qquad (18)$$

$$y_k = [o_m \ 1]\begin{bmatrix} \Delta\tilde{x}_k \\ \tilde{y}_k \end{bmatrix} \qquad (19)$$

Where:
- $o_m = [0 \ 0 \ 0 \ ... \ 0] \in R^{1 \times m}$, $m$ is the number of state variables.

The following definitions must be introduced to design the model predictive control:
- control horizon: number of predicted input sets, as:
$$(\Delta u_k, \Delta u_{k+1}, \Delta u_{k+2}, ..., \Delta u_{k+N_c-1})$$
where $N_c$ is the length of the control horizon,

- prediction horizon: number of predicted system states, as:
$$(\Delta x_{k+1,k}, \Delta x_{k+2,k}, \Delta x_{k+3,k}, ..., \Delta x_{k+N_p,k})$$
where $N_P$ is the length of the prediction horizon.

Using the augmented model equations from (15) – (19) the predicted input sets and system states can be defined. This is written in (20) and (21).

$$x_{k+1,k} = Ax_k + B\Delta u_k$$
$$x_{k+2,k} = A^2 x_k + AB\Delta u_k + B\Delta u_{k+1}$$
$$\vdots$$
$$x_{k+N_p,k} = A^{N_p}x_k + A^{N_p-1}B\Delta u_k + A^{N_p-2}B\Delta u_{k+1} ... + A^{N_p-N_c}B\Delta u_{k+N_c-1} \qquad (20)$$

$$y_{k+1,k} = CAx_k + CB\Delta u_k$$
$$y_{k+2,k} = CA^2 x_k + CAB\Delta u_k + CB\Delta u_{k+1}$$
$$\vdots$$
$$y_{k+N_p,k} = CA^{N_p}x_k + CA^{N_p-1}B\Delta u_k + \cdots + CA^{N_p-N_c}B\Delta u_{k+N_c-1} \qquad (21)$$

From equations (20) and (21) a new state equation can be introduced as defined in (22).

$$Y = Fx_k + S\Delta U \qquad (22)$$

Where:

$$Y = [y_{k+1,k} \ y_{k+2,k} \ \cdots \ y_{k+N_p,k}]^T \qquad (23)$$

$$\Delta U = [\Delta u_k \ \Delta u_{k+1} \ ... \ \Delta u_{k+N_c-1}]^T \qquad (24)$$

$$F = \begin{bmatrix} CA \\ CA^2 \\ \vdots \\ CA^{N_p} \end{bmatrix} \qquad (25)$$

$$S = \begin{bmatrix} CB & 0 & \cdots & 0 \\ CAB & CB & \ddots & 0 \\ \vdots & \vdots & \ddots & \vdots \\ CA^{N_p-1}B & CA^{N_p-2}B & \cdots & CA^{N_p-N_c}B \end{bmatrix} \qquad (26)$$

The output reference vector must be given. This vector contains all target values for the output variables. In this case, this means target values for vehicle position, velocity and heading angle. The form of the reference vector is defined in (27).

$$R_s^T = \begin{bmatrix} I^{K \times K} \ r_k \\ \vdots \end{bmatrix} \in R^{N_p K \times K} \qquad (27)$$

Where:
- $K$ is the number of outputs,

- $r_k$ is a column vector containing the reference values of the output variables in $k^{th}$ time step.

The cost function of the control problem now can be formed according to (28). After the cost function is defined, the optimum input set is searched where the cost is minimal. This can be done by deriving the cost function according to the input increment set. The derivation step is not detailed in the current paper but can be found in [4].

$$J = (R_s - Y)^T(R_s - Y) + \Delta U^T \bar{R} \Delta U \quad (28)$$

The cost terms are the output deviation from their reference values and the amplitude of the input signals. The optimal input increment set is defined in (29).

$$\Delta U = (S^T S + \bar{R})^{-1} S^T (R_s^T - F x_k) \quad (29)$$

Where:
- $\Delta U$ is the optimal input increment vector for the control horizon,
- $\bar{R}$ is a diagonal matrix containing the weight factor. This factor describes the weight of the amplitude of the inputs in the cost function. Higher weight results less aggressive control behavior.

### B. Constrained model predictive control

Equation (29) is a solution of the cost optimization problem without any constraints. In practice, intervention signals are often limited by actuator saturations (e.g. maximum steering angle) or by comfort requirements (e.g. maximum lateral acceleration), or by any other physical considerations. The optimization of (28) can be formed to a general quadratic optimization problem as written in (30) and (31).

$$J = \frac{1}{2} x^T E x + x^T F$$

$$subj: Mx \leq \gamma \quad (30)$$

Equation (30) is the general formula of a quadratic problem which must be solved subjected to the equality and inequality constraints. The optimization formula within the MPC can be written as in (31).

$$x = \Delta U$$
$$E = 2\left(S^T S + r_w I_{N_c \times N_c}\right)$$
$$F = -2\left(S^T(\bar{R}_s r_k - F x_a)\right) \quad (31)$$

A quadratic problem can be solved with many different approaches. In our solution we used Hildreth's iterative solver with a maximum of 38 iterations. The theoretical background is described in [4].

### C. Implementation for vehicle control

The kinematic model presented in (7) – (14) is used for the MPC design. For two different simulation environments two different input vectors are used. The kinematic model from Simulink expects a road wheel angle and longitudinal acceleration reference. The LGSVL model expects the road wheel angle as well, however the longitudinal control expects velocity reference instead of the acceleration. The velocity control is handled within the model in this case. Due to this difference the controller should not be modified, as the velocity reference can be directly forwarded through the controller and can be connected to the low-level speed controller of LGSVL. As written in (13) the outputs of the model are the lateral displacement, orientation and the longitudinal speed. The reference values for these outputs are provided by a superordinate planner function. These reference values are represented by a waypoint array, which is resampled within the controller to have smoother behavior.

The controller was realized in Simulink and was also directly built as a Robot Operating System (ROS) node with the help of Simulink Coder.

## V. RESULTS

In this chapter the performance of the MPC function introduced in previous chapters is presented. These results have been generated using a pre-defined trajectory. This trajectory covers a simple side displacement maneuver in an S-like curve. This can be a basis for a lateral parking maneuver. The vehicle starts from standstill, accelerates to a constant speed and decelerates to complete stop. Driving direction is forward during the entire maneuver. The path run by the vehicle is examined. The controller performance is compared with a kinematic bicycle model implemented in Simulink and a complete vehicle dynamic model in LGSVL simulation environment. It must be emphasized that the model used for the MPC only considers vehicle kinematics and does not include neither the dynamic effects of the vehicle nor the dynamics of the steering system. In addition to the vehicle paths the steering angle and the velocity control characteristics are shown.

In Fig. 2 the vehicle path can be seen for two different simulation: one with the kinematic bicycle model implemented in Simulink, and the other with LGSVL simulation environment. Both simulations have been triggered by exactly the same trajectory points for both lateral and longitudinal target values.

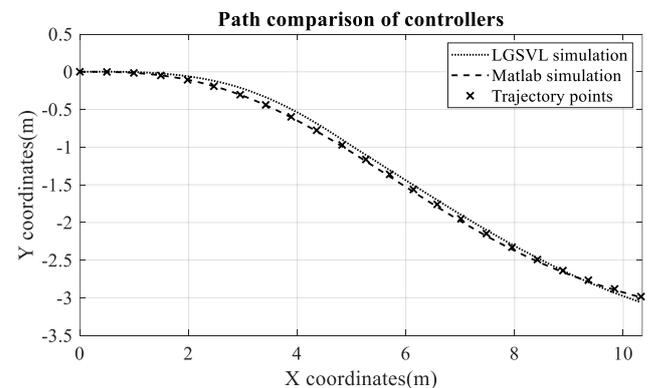

Fig. 2. Path of the vehicle in two different simulation environments

As seen on Fig. 2, the vehicle follows the trajectory as intended. The result from the LGSVL simulation indicates that the vehicle slightly overshoots the first turn even after calibrating the controller. This kind of behavior was observable for different parameter sets, which assumes a systematic malfunction of the controller. In Fig. 3 the target and the actual steering angle can be seen. With the kinematic bicycle Simulink model. The target and the actual value are smooth. The small deviation between them is caused by an artificial delay element modelling the discrete characteristics of the system. In case of the LGSVL simulation a strong oscillation can be observed. This is most probably caused by the dynamics of the steering. These dynamics are not considered when designing the controller. This must be further analyzed. A feasible solution may be to add the steering model to the controller. The high frequency noise on the steering command from LGSVL is caused by a communication malfunction between the simulator and the ROS environment. This effect can be eliminated by reducing the controller time stamp, however that would cause delay in the control loop.

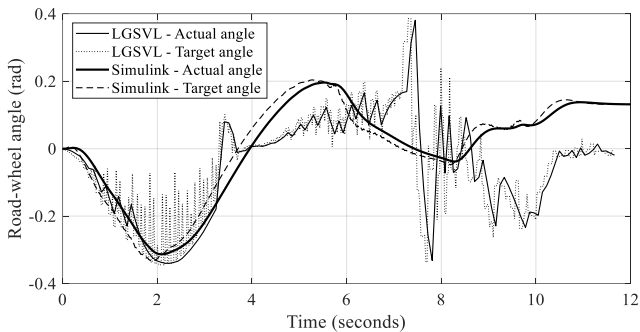

Fig. 3. Steering signals during the maneuver – the maneuver in the Matlab simulation took longer time as the velocity acceleration of the vehicle is smaller than in LGSVL.

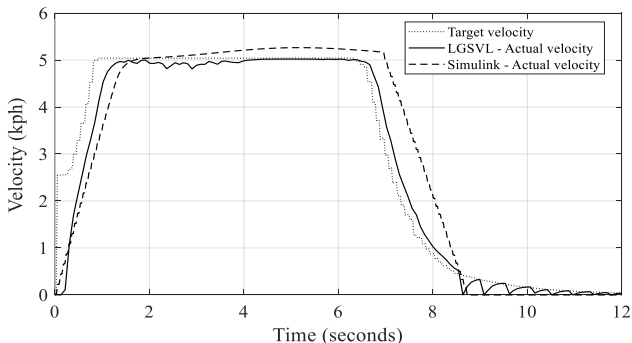

Fig. 4. Speed signals during the maneuver

In Fig. 4 the speed signals can be seen. This figure is only shown for informative reasons. The comparison between the two environments would not be meaningful due to the fact, that different reference values are provided by the controller for the two environments. However, based on this figure the maneuver can be reconstructed: starting off from standstill, maintaining a relatively low speed and finishing with a full stop.

## VI. SUMMARY AND FUTURE WORK

Based on the results the controller concept is proven to work. However, there is a deviation in the controller performance when using it with a complete vehicle model. The dynamics of the vehicle at such a low speed are not significant, therefore a kinematic model used for the controller is satisfactory. However, lacking the model of the steering can lead to controller performance issues. Even though the effect of neglecting the steering model is acceptable on the vehicle position level, it can lead to serious jerking steering which is not desired. Therefore, our next step would be to examine the transfer characteristics of the steering and extend the vehicle model with the steering mechanics. It is expected to have much smoother steering behavior, as seen in the case of the Simulink model simulations. Also analysis of steering jerks at speed close to zero is necessary.

We intend to execute real-vehicle tests. The controller is already integrated as a Robot Operating System (ROS) node into a higher level autonomous system. As the same infrastructure is used in a real vehicle than that of the LGSVL simulation, vehicle test will be easy to be accomplished.

A wider scope of simulations is planned to be run in the near future to test the operating range of the controller using the kinematic vehicle model. It is expected to have an upper speed limit of this model. During these tests we are planning to make comparison analysis between MPC and other vehicle control schemes, such us Pure-pursuit or Stanley.

## VII. ACKNOWLEDGMENT

The research presented in this paper was carried out as part of the "Autonomous Vehicle Systems Research related to the Autonomous Vehicle Proving Ground of Zalaegerszeg (EFOP-3.6.2-16-2017-00002)" project in the framework of the New Széchenyi Plan. The completion of this project is funded by the European Union and co-financed by the European Social Fund.